# Relative Facial Action Unit Detection

Mahmoud Khademi
Department of Computing and Software
McMaster University
Hamilton, Canada
khademm@mcmaster.ca

Louis-Philippe Morency
Institute for Creative Technologies
University of Southern California
Los Angeles, USA
morency@ict.usc.edu

*Abstract*—This paper presents a subject-independent facial action unit (AU) detection method by introducing the concept of relative AU detection, for scenarios where the neutral face is not provided. We propose a new classification objective function which analyzes the temporal neighborhood of the current frame to decide if the expression recently increased, decreased or showed no change. This approach is a significant change from the conventional *absolute* method which decides about AU classification using the current frame, without an explicit comparison with its neighboring frames. Our proposed method improves robustness to individual differences such as face scale and shape, age-related wrinkles, and transitions among expressions (e.g., lower intensity of expressions). Our experiments on three publicly available datasets (Extended Cohn-Kanade (CK+), Bosphorus, and DISFA databases) show significant improvement of our approach over conventional absolute techniques.

## I. Introduction

During face-to-face communication people naturally exchange information through their verbal and nonverbal behaviors. Facial expressions are an important part of natural human communication and new technologies for human-computer/robot interaction and intelligent environments are taking advantages of this communicative channel. Automatic facial expression analysis can help making the interaction between human and computer more flexible and robust. Recent application areas include pain assessment, psychological science, and neurology [1, 2].

Over the last decade, there have been a lot of attempts to perform the task of automatic analysis of facial expressions [2]. However, developing an automatic real-time facial expression analysis system that can deal with pose variations, illumination variations, occlusion, head motions, lower intensity of expressions, and individual differences across subjects is still a challenging task in computer vision and machine learning communities. At the center of automatic facial expression analysis is challenge of detecting the movements of some individual facial muscles called facial action units (AUs).

Conventional AU detection techniques follow an absolute or frame-based approach where the classification is performed based on the current frame, without explicit comparison with neighboring frames. Given their absolute nature, these frame-based approaches have a hard time to handle scenarios with different levels of expression intensity and individual differences across subjects. For instance, as shown in Fig. 1, an actor can naturally show some permanent furrows and wrinkles around the eyes, even during a neutral face. These age-related wrinkles, may interpret by the classifier as presence of AU4 in this frame. Frame-based methods will likely to detect the presence of AU4 (frowning) in this case since the eyebrows of this actor are to some extent lower, even in the neutral face.

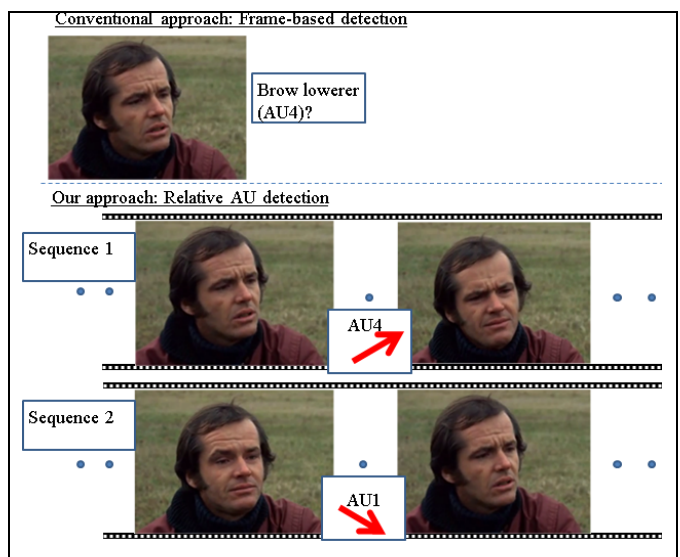

Figure 1. Frame-based versus relative facial AU detection. A frame-based method is more likely to decide the presence of facial AU4 incorrectly in the first image of sequence 1 (and second image of sequence 2). However, a relative comparison between adjacent frames simplifies the problem of AU identification for both sequences 1 and 2.

In this paper, we propose a new method for facial AU detection which analyzes the temporal neighborhood of the current frame to decide if the expression recently increased, decreased or showed no change. By focusing on the relative changes in shape and texture of the face, our approach can improve robustness to individual differences such as face scale and shape, age-related wrinkles, and transitions among expressions (e.g., lower intensity of expressions). Our new classification objective function is more robust to individual differences across subjects and different levels of expression. As shown in Fig. 1, a relative comparison between adjacent frames simplifies the problem of AU identification for both sequences 1 and 2.

The rest of the paper has been organized as follows: In section II, we review the related works. Section III discusses our proposed method for detecting relative changes of a facial

AU in an input video. In section IV, we first describe the method which we used for representation of facial AUs by applying geometric and appearance facial features. Then, we report our experimental results. Section V presents conclusions and future research directions.

## II. RELATED WORKS

Several survey papers reviewed the facial expression analysis researches [2, 3]. To code facial AUs, Ekman and Friesen developed Facial Action Coding System (FACS) [4]. The FACS includes 44 facial AUs. Among them 30 AUs are related to movement of a particular set of muscles: 12 for upper face and 18 for lower face [5].

An automatic facial expression analysis system typically has facial feature extraction and classification stages. There are two major types of facial features: geometric and appearance features. The geometric facial features give the shape and locations of the face parts. Appearance-based features can be extracted by applying wavelets like Gabor [6] to the whole or specific locations of the face. While local methods are suitable for subtle change in small locations, holistic approaches are good at representing common facial expressions. More recently, methods based on combinations of geometric and appearance features are common. Senechal et al. [7] used Active Appearance Model (AAM) coefficients and Local Gabor Binary Pattern (LGBP) histograms to detect facial AUs. Simon et al. [8] applied AAM with Scale-Invariant Feature Transform (SIFT) descriptors at some landmark points on face images.

Facial AU and expression classification methods can also be divided in two major groups: frame-based and sequence-based recognition methods. The frame-based methods use only a single frame with or without the neutral face. The sequence-based methods, however, use the time-related (temporal) information of the sequence. It is important to note that although these time-related approaches are using temporal information, they are not explicitly comparing neighboring frames but instead are learning dynamic between labels or latent variables. For example, methods like Hidden Markov Models (HMMs) [9] and rule-based classifier [10] have been used.

The state-of-the-art techniques use facial expression's temporal dynamics for sequence-based classification of AUs. Valstar and Pantic [11] used temporal information by analyzing the phase of the facial AU, i.e. onset, apex, or offset and the activation duration. Another notable work by Zhu et al. [12] is applying dynamic cascades with bi-directional bootstrapping to select an optimal training set.

These previous methods did not explicitly address the problem of high individual differences among subjects such as face scale, permanent furrows as well as lower intensity of expressions. Such individual differences in subjects may lead to poor classification results especially when the training set is small. Another difficulty is transitions among expressions. Transitions from AUs to another may contain no intervening neutral state. In other words, an expression may not start and end with neutral position. To recognize these AUs, the training data must contain dynamic combinations of AUs. We will address these problems by introducing relative facial AU detection method.

## III. RELATIVE FACIAL ACTION UNIT DETECTION

We define the relative facial AU detection problem as follows: given a frame $I_t$ from an input image sequence and a specific AU to estimated, our objective is to determine if the intensity of the AU in $I_t$ is increasing, decreasing, or there is no change. In this section, we first describe our generic relative facial AU detection approach based on pair-wise comparison of the neighborhood frames. Then, we discuss how to train the proposed system using a set of image pairs.

### A. Overview

Fig. 2 gives an overview of our method to assign the relative frame labels by incorporating neighboring information. When analyzing the expression intensities over time, the exact time of the onset, apex, and offset of facial expression is often fuzzy. As a result, computing the relative labels is more likely to be accurate than a method which uses only the current frame.

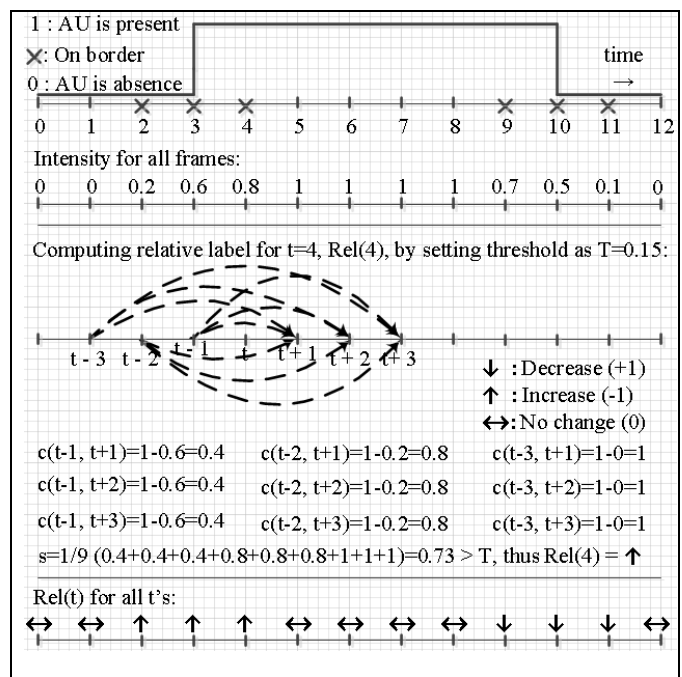

Figure 2. Top: The intensity of the AU for each frame. Middle: computing the relative label for t = 4. We used $3 \times 3 = 9$ comparisons between previous and next frames. The value $c(t, \tilde{t})$ is the output of a classifier for image pair $(I_t, I_{\tilde{t}})$, which can be interpreted as the difference between intensity and motion changes of the image pair. Then, we computed the average of the relative values (classifier outputs). Bottom: relative class labels for all frames of the sequence.

The first step of our approach is to train a pair-wise classifier which differentiate between increased, decreased or no change (see Section III.B for more details). The output of this classifier can be interpreted as the difference between intensity and motion changes of the image pairs. During testing, we estimate the relative labels using the neighboring frames. The neighborhood is defined by window size. In the example shown in Fig. 2, we used a window of size $\mp 3$, which represents a total neighborhood of 7, including the current frame. Given this neighborhood, we compare all prior frames with all next frames. In the example, $3 \times 3 = 9$ comparisons were performed. For each comparison, a relative labels is estimated using our pair-wise classifier. Then, we aggregate the relative labels to predict the final relative label of the current frame. Formally, the relative label of the current frame $I_t$ can be defined as follows:

$$Rel(t) = \begin{cases} \uparrow & if & s > T \\ \downarrow & if & s < -T \\ \leftrightarrow & if & |s| < T \end{cases} \quad (1)$$

Where $s = 4/w^2 \sum_{i=1}^{w/2} \sum_{j=1}^{w/2} c(t-i, t+j)$, $c(t, \tilde{t})$ is the output of the classifier for image pair $(I_t, I_{\tilde{t}})$, which is a real value between $-1$ to $1$, $T$ is a threshold and w is the window size. We divided the sum by $w^2/4$, since we must take the average of $(w/2) \times (w/2)$ comparisons. Ideally, $c(t, \tilde{t})$ is the subtraction of the intensity of AU in the first image from the second one. The change in the intensity of the AU in images $I_t$ and $I_{\tilde{t}}$ is negligible, if the absolute of the average of the relative values,$|s|$, is less than the threshold $T$. The threshold and window size are hyper-parameters of the algorithm and are automatically determined during validation.

### B. Pair-wise Classifier

Given the facial expression intensity (i.e., continuous output labels) of two frames, we train a three-class classifier which assigns labels +1, -1, or 0 to represent increase, decrease or no change. Given a specific AU and two frames of a subject, we assign label -1 (decrease) to the image pair, if the AU is present in the first image but it becomes absence in the second image. Similarly, label +1 (increase) will assign to the image pair when the AU is absence in the first image but it becomes present in the second image. When AU is absence or present in both frames, we assign label 0 (no change) to the image pair. In this non-symmetric three-class pattern classification problem the distribution of the third class is more diverse than other classes.

Each image pair $(I_t, I_{\tilde{t}})$ can be either two static images of a subject or consecutive frames of an annotated image sequence. In Fig. 2, for example, $(I_1, I_5)$ and $(I_1, I_6)$ can be selected as training samples with label +1. Similarly, $(I_7, I_{12})$ and $(I_8, I_{12})$ are good candidates with label -1. $(I_0, I_1)$ and $(I_5, I_7)$ are also two samples with label 0. The frames which are on border are not good for the purpose of training since their presence/absence labels are noisy, especially if the intensity is not provided.

## IV. EXPERIMENTAL RESULTS

### A. Experimental Setup

*1) Geometric-Based Facial Feature Extraction:* We used the 3D Constrained Local Model (CLM-Z) [13] to track 66 facial feature points in the consecutive frames. The CLM-Z is a robust facial feature tracker. We combined the Generalized Adaptive View-based Appearance Model (GAVAM) [14], a rigid head pose tracker, with CLM-Z to eliminate rigid head motions from non-rigid facial movements (see Fig. 3). Given frame $I_t$, GAVAM estimates the pose vector $x_t$, by a 6-dimensional vector composed of the 3D translation and three Euler angles corresponding to pitch, yaw, and roll: $x_t = [r_x, r_y, r_z, \omega_x, \omega_y, \omega_z]$.

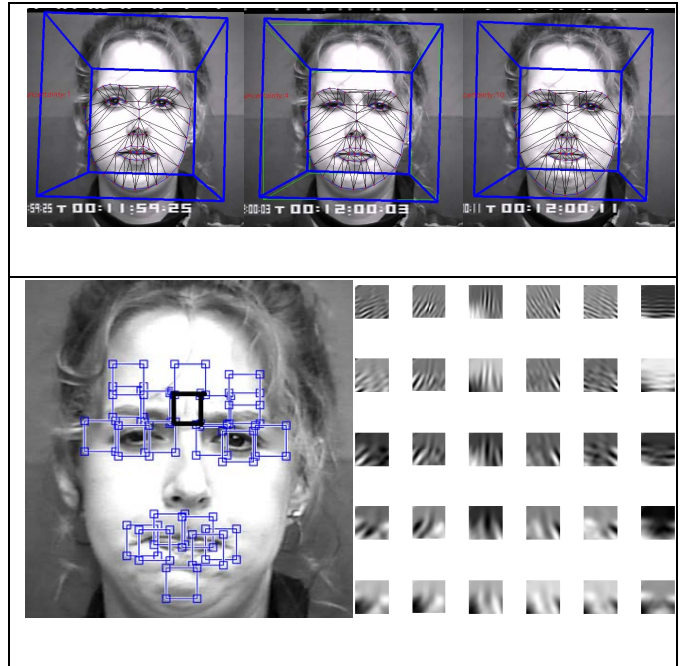

Figure 3. Facial feature extraction. Top: geometric-based facial feature extraction by a facial feature point tracker called 3D Constrained Local Model (CLM-Z) and a rigid head pose tracker called Generalized Adaptive View-based Appearance Model (GAVAM). Bottom: appearance-based facial feature extraction by applying Gabor wavelets to some patches of the face. Different Gabor responses resulted from convolving an image patch (the region between eyebrows) with real part of the Gabor wavelets (with 5 scales and 6 orientations) are showed at right of the face.

Let $p_i = (X, Y, Z)$ denotes the 3D location of the $i'$th facial feature point computed by CLM-Z in the GAVAM frame of reference. We can simply wrap the shape by the following equation:

$$\bar{p}_i = (p_i - r) \times R_z(-\omega_z) \times R_y(-\omega_y) \times R_x(-\omega_x) \quad (2)$$

Where $\bar{p}_i$ is the 3D location of the $i'$th facial feature point after eliminating the translation and rotation due to rigid head motions, $r = [r_x, r_y, r_z]$ is the 3D translation, $R_z, R_y$ and $R_x$ are the 3D rotation matrices which rotate vectors around the $x, y,$ and $z$ axis with Euler angles $\omega_z, \omega_y$ and $\omega_x$ respectively.

In this full perspective model, we can compute the coordinate x and y of the $i$'th feature point in the image plane by following equations:

$$x = \frac{fX}{Z} + c_x \quad (3)$$

$$y = \frac{fY}{Z} + c_y \quad (4)$$

Where $f$ is the camera focal length, and $c_x, c_y$ are camera central points. We used the distance between specific facial feature points as the geometric features of frame $I_t$. The geometric feature vector $g$ for an image pair $(I_t, I_{\tilde{t}})$ is the concatenation of the geometric features in frames $I_t$ and $I_{\tilde{t}}$. That is, $g = (g_t, g_{\tilde{t}})$.

*2) Appearance-based Facial Feature Extraction:* For each AU we can select a number of image patches, which are subject to change by presence of that AU, to apply an image filters. For example, for AU4 (brow lowerer) a suitable candidate is the region between two eyebrows which are usually subject to changes by wrinkles and furrows during presence of AU4. The Appearance-based feature extraction, were made based on the state-of-the-art method describe in [7] where they reported best results with LGBP histograms and geometric features.

Before applying the filters, the texture must be wrapped. We used a specific set of facial feature points to find the image patch in each frame. Then, we split the patch in two triangles and wrapped the texture using barycentric coordinate system and the pose vector $x_t$ resulted from GAVAM.

To extract the appearance-based facial features from each frame, we used a set of Gabor wavelets [6] with different scales and orientations. The real and imaginary pats of the Gabor function are as follows:

$$f_1(\boldsymbol{u}) = \frac{|\boldsymbol{k}|^2}{\sigma^2} \exp\left(\frac{|\boldsymbol{u}|^2|\boldsymbol{k}|^2}{-2\sigma^2}\right) \times$$
$$(\cos(\boldsymbol{k}.\boldsymbol{u}) - \exp\left(\frac{-\sigma^2}{2}\right)) \quad (5)$$

$$f_2(\boldsymbol{u}) = \frac{|\boldsymbol{k}|^2}{\sigma^2} \exp\left(\frac{|\boldsymbol{u}|^2|\boldsymbol{k}|^2}{-2\sigma^2}\right) \times$$
$$(\sin(\boldsymbol{k}.\boldsymbol{u}) \quad (6)$$

where $\boldsymbol{u} = (x, y)$, is the variable in spatial domain and $\boldsymbol{k} = (\beta\cos(\theta), \beta\sin(\theta))$ is the frequency vector which defines the scale $\beta$ and direction $\theta$. We used 5 scales and 6 orientations. The parameter $\sigma$ is also set to $\pi$. By convolving real and imaginary parts $f_1$ and $f_2$ with each image patch and computing the Gabor magnitude, we can represent the patch with $5 \times 6 = 30$ Gabor magnitudes. Fig. 3 represents 30 Gabor responses by applying the real part $f_1$ to a region between eyebrows.

Then, we applied the Local Binary Pattern operator to 30 Gabor magnitudes and computed LGBP histograms with 256 bins for them (for more details see [7]). By concatenating all histograms, we can represent each image by $30 \times n \times 256$ features, where n is the number of patches which depend on the AU. Then, we subtracted the feature vector of the first image from the second one. That is, $\boldsymbol{a} = \boldsymbol{a}_{\tilde{t}} - \boldsymbol{a}_t$. In this way, we can represent each image pair $(I_t, I_{\tilde{t}})$ by a vector of dimension $30 \times n \times 256$. Finally, we applied isometric feature mapping (Isomap) algorithm [15], which is a manifold learning algorithm, to the resulted vectors to reduce the dimensionality and remove the dependencies between appearance features. The target dimensionality in Isomap algorithm was set to 40.

*3) Classification:* We may concatenate the geometric and appearance feature vectors to classify the samples. However, since the geometric and appearance features are basically different [16], we used a more efficient method by applying Kernel Canonical Correlation Analysis (KCCA) algorithm to two different views of the data proposed by Meng et al. [17, 18]. We used a three-class SVMs with two RBF kernels with different parameters for two views to classify the image pairs. When the intensity of AUs is provided, a classifier with continuous labels may be trained. We used LIBSVM library with the epsilon-SVR option (see [19] for definition of LIBSVM options and parameters). The best penalty parameter $C$ and two kernel parameters $\gamma_1$ and $\gamma_2$ were obtained by searching in a wide range of parameter values using a small subset of data. The best model was selected based on $F1$ score:

$$F1 = 2 \frac{\text{precision}.\text{recall}}{\text{precision} + \text{recall}} \quad (7)$$

*4) Baseline:* In order to compare our method with frame-based facial AU detection approach, we implemented a baseline. In both baseline and proposed methods, we design a classifier for each single AU. Having the presence/absence labels for both images $I_t$ and $I_{\tilde{t}}$, we may obtain the class label of the image pair $(I_t, I_{\tilde{t}})$. When the intensity of AUs is not provided, we considered the discrete case. That is, for each single image $I_t$, if the AU is present then Intensity$(I_t) = 1$, otherwise Intensity$(I_t) = 0$. When the intensity of AUs is provided, we used 5 intensity levels 0.580, 0.685, 0.790, 0.895, and 1.0 corresponding to the intensity levels A, B, C, D, and E in FACS. With this notation we can define the target for an image pair $(I_t, I_{\tilde{t}})$ as follows:

$$c(t, \tilde{t}) = \text{Intensity}(I_{\tilde{t}}) - \text{Intensity}(I_t) \quad (8)$$

In the baseline method, the geometric feature vectors $\boldsymbol{g}_t$ and appearance feature vector $\boldsymbol{a}_t$ were extracted from image $I_t$. Similarly, the geometric feature vectors $\boldsymbol{g}_{\tilde{t}}$ and appearance feature vector $\boldsymbol{a}_{\tilde{t}}$ were extracted from image $I_{\tilde{t}}$. For both baseline and proposed methods the same dimensionality reduction algorithm was applied to reduce the appearance features. Then, we applied KCCA to $(\boldsymbol{g}_t, \boldsymbol{a}_t)$ and $(\boldsymbol{g}_{\tilde{t}}, \boldsymbol{a}_{\tilde{t}})$ separately. Finally, a two-class SVMs was trained to decide about presence/absence of the AU in images $I_t$ and $I_{\tilde{t}}$. When intensity of AUs is provided, LIBSVM library with the epsilon-SVR option was used. In this case, the output of the baseline classifier is a value between 0 to 1 representing the intensity of the AU.

*B. Datasets*

We tested our algorithm using three databases: Extended Cohn-Kanade (CK+) database [20, 21], Bosphorus database [22], and Denver Intensity of Spontaneous Facial Action (DISFA) database [23, 24].

*1) CK+ Database:* There are 582 image sequences of 123 subjects in CK+ database. The pick frame for each sequence is FACS coded. A benchmark is also provided. For each subject, we chose the peak frame of every expression together with the neutral face. Then, we created several image pairs by selected frames. For most of the image pairs in the training set, one of the frames is neutral face. However, for some image pairs both frames represent the presence of different AUs. This enables the algorithm to detect AUs which are involving no intervening neutral state. Given an AU we can determine the class label for the image pairs by (8). In this way, 1164 image pairs were created for all 123 subjects.

*2) Bosphorus Database:* There are 105 subjects and 4666 static images in Bosphorus database. Each image is FACS coded. The intensity for each AU is also available. The database includes 3D and 2D coordinates of 24 facial feature points for each static image (for this database we did not apply CLM-Z). There are up to 35 expressions per subject. We created 2790 image pairs using this database.

*3) DISFA Databese:* To test the proposed method by real-world data, we evaluated the performance of the proposed and baseline methods using videos of DISFA database. DISFA database includes spontaneous AUs of 27 adult subjects, in which each subject has been video recorded for 4 minutes (i.e. 20 frames per second). Every frames of this database is coded by a FACS coder. The intensity of AUs is also provided.

### C. Methodology

For CK+ and Bosphorus databases, the test was performed using Leave-one-subject-out cross-validation method. For DISFA database, we also used image pairs of CK+ and Bosphorus databases to train the classifier. More precisely, we extracted a set of around 4000 image pairs from 26 subjects of DISFA database using the procedure described in section III. Then, we added 3912 samples (image pairs) of CK+ and Bosphorus databases to this set and trained a SVMs classifier. The remained subject was used to test the classifier accuracy. For this purpose, we defined the relative value for each frame of the remained video as described in Fig. 2. A window of size 10 was used for the proposed method using a small subset of training data. Also, the threshold in (1) was set to 0.15.

### D. Results and Discussion

Table 1 shows the $F1$ scores of upper and lower face AUs for the baseline and proposed methods on CK+ and Bosphorus databases. Experiments by the relative method show improvement in terms of $F1$ score in comparison to the frame-based baseline method. For CK+ database, the average $F1$ score is 0.790 and 0.816 for the baseline and the relative methods, respectively. For Bosphorus database, the average $F1$ score is 0.683 and 0.729 for the baseline and the relative methods, respectively. We also did the pairwise t-test. As we can see from the $p$ value of the t-test, the proposed method shows statistically significant improvement in terms of $F1$ score.

To compare the proposed method with a common benchmark [20], we designed another experiment. Since the geometric feature extraction of our method is the same as the benchmark results on CK+, we only used the geometric features for this experiment. We first applied the proposed methods to a set of image pairs of CK+ database and obtained the relative labels. Then, we decided about the presence/absence of AU in both frames based on the relative labels. Due to compatibility with the benchmark results we used the area underneath the ROC curve as the measure in this experiment. Table 2 shows the area underneath the ROC curve for the proposed and the benchmark results. The average of the area underneath the ROC curve for the proposed method and benchmark results is 0.915 and 0.899, respectively.

TABLE 1. EXPERIMENTAL RESULTS ON CK+ AND BOSPHORUS DATABASES

| AUs | F1 score for relative labels on CK+ | | F1 score for relative labels on Bosphorus | |
|---|---|---|---|---|
| | Baseline | Proposed (Relative) | Baseline | Proposed (Relative) |
| AU1 | 0.81 | 0.82 | 0.63 | 0.66 |
| AU2 | 0.85 | 0.87 | 0.56 | 0.60 |
| AU4 | 0.77 | 0.81 | 0.64 | 0.75 |
| AU5 | 0.76 | 0.81 | 0.52 | 0.56 |
| AU6 | 0.67 | 0.70 | - | - |
| AU7 | 0.64 | 0.67 | - | - |
| AU9 | 0.87 | 0.91 | 0.76 | 0.84 |
| AU12 | 0.84 | 0.88 | 0.85 | 0.85 |
| AU15 | 0.76 | 0.78 | 0.65 | 0.74 |
| AU16 | - | - | 0.60 | 0.64 |
| AU17 | 0.80 | 0.82 | 0.61 | 0.64 |
| AU20 | 0.73 | 0.76 | 0.67 | 0.68 |
| AU22 | - | - | 0.78 | 0.84 |
| AU23 | 0.75 | 0.76 | 0.54 | 0.60 |
| AU25 | 0.88 | 0.90 | 0.83 | 0.87 |
| AU26 | - | - | 0.70 | 0.75 |
| AU27 | 0.93 | 0.94 | 0.90 | 0.91 |
| **Average** | **0.790** | **0.816** | **0.683** | **0.729** |
| **Variance** | **0.007** | **0.006** | **0.014** | **0.013** |
| **p value of t-test** | $3.2 \times 10^{-6}$ | | $4.2 \times 10^{-5}$ | |

TABLE 2. EXPERIMENTAL RESULTS ON CK+ DATABASE (FOR PRESENCE/ABSENCE LABELS) AND DISFA DATABASE

| AUs | Area underneath the ROC curve for presence/absence labels on CK+ | | Accuracy for relative labels on DISFA | |
|---|---|---|---|---|
| | Benchmark [20] | Proposed (Relative) | Baseline | Proposed (Relative) |
| AU1 | 0.94 | 0.95 | 0.82 | 0.84 |
| AU2 | 0.97 | 0.97 | 0.83 | 0.86 |
| AU4 | 0.86 | 0.89 | 0.78 | 0.82 |
| AU5 | 0.95 | 0.97 | 0.88 | 0.92 |
| AU6 | 0.92 | 0.94 | 0.78 | 0.80 |
| AU7 | 0.78 | 0.81 | - | - |
| AU9 | 0.98 | 0.98 | 0.87 | 0.89 |
| AU12 | 0.91 | 0.93 | 0.74 | 0.77 |
| AU15 | 0.80 | 0.83 | 0.86 | 0.88 |
| AU17 | 0.84 | 0.86 | 0.73 | 0.75 |
| AU20 | 0.91 | 0.93 | 0.87 | 0.88 |
| AU23 | 0.91 | 0.92 | - | - |
| AU25 | 0.97 | 0.97 | 0.72 | 0.74 |
| AU26 | 0.75 | 0.77 | 0.75 | 0.76 |
| AU27 | 1.00 | 1.00 | - | - |
| **Average** | **0.899** | **0.915** | **0.802** | **0.826** |
| **Variance** | **0.006** | **0.005** | **0.004** | **0.004** |
| **p value of t-test** | $1.2 \times 10^{-4}$ | | $5.1 \times 10^{-6}$ | |

Table 2 shows the experimental results on DISFA dataset too. Since there are different levels of intensity, we use the accuracy as the measure for this experiment. The average accuracy is 0.802 for the baseline and 0.826 for the relative method. As we can see from the p value of the t-test, the proposed method shows statistically significant improvement for real-world videos of DISFA database too.

Although better results may be achieved by using more efficient appearance features, we can see the proof of relative facial AU detection concept, since we used the same features and classifier for both approaches. Moreover, applying our approach on a new test sequence, including all tracking, image filters, dimensionality reduction and classification computations, can be performed in less than 0.03 seconds per frame with moderate computing power. As a result, the system is suitable for real-time applications.

## V. Conclusion and Future Research Directions

We proposed an efficient subject-independent facial AU detection method using classification of relative change in facial AUs, when the neutral face is not provided. In contrast to the frame-based facial AU detection methods which use only a single frame, our relative AU detection approach uses previous and next frames to decide about increase, decrease, or no change. Experiments show the proposed system is more robust to individual differences among subjects, transitions among expressions and lower intensity of expressions.

Our relative approach can easily be extended to temporal models such as Conditional Random Fields (CRFs) and HMMs, since the main contribution of this paper is centered around the new relative feature descriptors which are independent on the classifier. As another future direction, we are interested to study the effect of other appearance-based feature extraction methods such as SIFT descriptors and Haar wavelets for relative facial AU detection method.

## References

[1] S. Z. Lee, A. K. Jain, "Handbook of Face Recognition," Springer Science+Business Media, Inc., ISBN 0-387-40595-X, New York, USA, 2004.

[2] M. Patnic, J. M. Rothkrantz, "Automatic analysis of facial expressions: The state of art," IEEE Transactions on PAMI, vol. 22, no. 12, December 2000.

[3] B. Fasel, J. Luettin, "Automatic facial expression analysis: a survey," Pattern Recognition, vol. 36, no. 1, pp. 259–275, 2003.

[4] P. Ekman, W. Friesen: The facial action coding system, "A technique for the measurment of facial movement," Consulting Psychologist Press, San Francisco, 1978.

[5] Y. Tian, T. Kanade, J.F. Cohn, "Recognizing action units for facial expression analysis," IEEE Transactions on PAMI, vol. 23, no. 2, 2001.

[6] M. Lyons, S. Akamatsu, M. Kamachi, J. Gyoba, "Coding facial expressions with Gabor wavelets," 3rd IEEE Int. Conf. on Automatic Face and Gesture Recognition, pp. 200–205, 1998.

[7] T. Senechal, V. Rapp, H. Salam, R. Seguier, K. Bailly, L. Prevost, "Combining AAM coefficients with LGBP histograms in the multi-kernel SVM framework to detect facial action units," IEEE International Conference on Automatic Face & Gesture Recognition and Workshops (FG 2011), pp. 860-865, 2011.

[8] T. Simon, M. H. Nguyen, F. De La Torre, and J. Cohn, "Action unit detection with segment-based svms," IEEE Conference on Computer Vision and Pattern Recognition (CVPR), pp. 2737–2744, 2010.

[9] I. Cohen, N. Sebe, F. Cozman, M. Cirelo, T. Huang, "Coding, analysis, interpretation, and recognition of facial expressions," Journal of Computer Vision and Image Understanding, Special Issue on Face Recognition, 2003.

[10] J. Cohn, T. Kanade, T. Moriyama, Z. Ambadar, J. Xiao, J. Gao, H. Imamura, "A comparative study of alternative faces coding algorithms," Technical Report CMU-RI-TR-02-06, Robotics Institute, Carnegie Mellon University, Pittsburgh, 2001.

[11] M. Valstar and M. Pantic, "Fully Automatic Facial Action Unit Detection and Temporal Analysis," Proceedings of the 2006 Conference on Computer Vision and Pattern Recognition Workshop (CVPRW'06), 2006.

[12] Y. Zhu, F. De la Torre, J. Cohn, and Y. Zhang, "Dynamic cascades with bidirectional bootstrapping for action unit detection in spontaneous facial behavior," IEEE Transactions on Affective Computing, vol. 2, issue 2, pp. 79-91, 2011.

[13] T. Baltrusaitis, P. Robinson and L.-P. Morency, "3D constrained local model for rigid and non-rigid facial tracking," Computer Vision and Pattern Recognition (CVPR), 2012.

[14] L.-P. Morency, J. Whitehill and J. Movellan, "Monocular head pose estimation using generalized adaptive view-based appearance model. Image and Vision Computing, doi:10.1016/j.imavis.2009.08.004, 2009.

[15] J. B. Tenenbaum, V. deSilva, and J. C. Langford, "A global geometric framework for nonlinear dimensionality reduction," Science, vol. 290, pp. 2319–23, 2000.

[16] B. Romera-Paredes., N. Bianchi-Berthouze, "Emotion recognition by two view SVM_2K classifier on dynamic facial expression features," IEEE Int. Conf. on Automatic Face & Gesture Recognition and Workshops (FG'11), pp. 854–859, 2011.

[17] H. Meng, D.R. Hardoon, J. Shawe-Taylor, and S. Szedmak, "Generic object recognition by combining distinct features in machine learning," vol. 5673, pp 90–98. SPIE, 2005.

[18] S. V. Vaerenbergh, "Kernel methods toolbox (KMBOX), a MATLAB toolbox for nonlinear signal processing and machine learning," http://sourceforge.net/p/kmbox, 2010.

[19] C.-C. Chang and C.-J. Lin, "LIBSVM: A Library for Support Vector Machines," ACM Trans. Intelligent Systems and Technology, vol. 2, no. 3, pp. 27, 2011.

[20] P. Lucey, J.F. Cohn, T. Kanade, J. Saragih, Z. Ambadar and I. Matthews, "The Extended Cohn-Kanade Dataset (CK+): A complete dataset for action unit and emotion-specified expression," Proceedings of IEEE workshop on CVPR for Human Communicative Behavior Analysis, San Francisco, USA, 2010.

[21] T. Kanade, J.F. Cohn, and Y. Tian, "Comprehensive database for facial expression analysis," Proceedings of the Fourth IEEE International Conference on Automatic Face and Gesture Recognition (FG'00), Grenoble, France, pp. 46-53, 2000.

[22] A. Savran, B. Sankur, M. T. Bilge, "Comparative evaluation of 3D versus 2D modality for automatic detection of facial action units," Pattern Recognition, vol. 45, pp. 767-782, 2012.

[23] S. M. Mavadati, M. H. Mahoor, K. Bartlett, and P. Trinh, "Automatic detection of non-posed facial action units," International Conference on Image Processing (ICIP), 2012.

[24] S. M. Mavadati, M. H. Mahoor, K. Bartlett, P. Trinh, and J. F. Cohn." DISFA: A spontaneous facial action intensity database," IEEE Transactions on Affective Computing, 2012